\RequirePackage{fix-cm}
\documentclass[a4paper,11pt]{article}

\usepackage{authblk}
\usepackage{algorithm}
\usepackage{algorithmic}
\usepackage{multirow}
\usepackage{graphicx}
\usepackage{booktabs}
\usepackage{epsfig}
\usepackage{epstopdf}
\usepackage{amsmath}
\usepackage{caption}
\usepackage{subcaption}
\usepackage{amsfonts}
\usepackage{wasysym}
\usepackage{paralist}
\usepackage{url}
\usepackage{float}
\usepackage{booktabs}
\usepackage{microtype}
\usepackage{tikz}
\usepackage{tikz-qtree}
\usepackage{adjustbox}

\newcommand{\F}{\mathbb{F}}

\newtheorem{lemma}{Lemma}

\providecommand{\keywords}[1]{\textbf{\textit{Keywords }} #1}

\begin{document}

\title{Evolving Constructions for Balanced, Highly Nonlinear Boolean Functions}

\author[1]{Claude Carlet}
\author[2]{Marko Djurasevic}
\author[2]{Domagoj Jakobovic}
\author[3]{Luca Mariot}
\author[3]{Stjepan Picek}
	
\affil[1]{{\small University of Bergen, Norway}

    {\small \texttt{claude.carlet@gmail.com}}}

\affil[2]{{\small Faculty of Electrical Engineering and Computing, University of Zagreb, Unska 3, Zagreb, Croatia}
	
	{\small \texttt{\{marko.durasevic,domagoj.jakobovic\}@fer.hr}}}

\affil[3]{{\small Digital Security Group, Radboud University, PO Box 9010, 6500 GL Nijmegen, The Netherlands} 
	
	{\small \texttt{\{luca.mariot,stjepan.picek\}@ru.nl}}}

\maketitle

\begin{abstract}
Finding balanced, highly nonlinear Boolean functions is a difficult problem where it is not known what nonlinearity values are possible to be reached in general. 
At the same time, evolutionary computation is successfully used to evolve specific Boolean function instances, but the approach cannot easily scale for larger Boolean function sizes. Indeed, while evolving smaller Boolean functions is almost trivial, larger sizes become increasingly difficult, and evolutionary algorithms perform suboptimally. In this work, we ask whether genetic programming (GP) can evolve constructions resulting in balanced Boolean functions with high nonlinearity. This question is especially interesting as there are only a few known such constructions. Our results show that GP can find constructions that generalize well, i.e., result in the required functions for multiple tested sizes. Further, we show that GP evolves many equivalent constructions under different syntactic representations. Interestingly, the simplest solution found by GP is a particular case of the well-known indirect sum construction.
\end{abstract}


\section{Introduction}
\label{sec:introduction}

Evolutionary algorithms (EAs) are successfully applied in various domains like rostering~\cite{Bai2010}, design of neural networks~\cite{9201169}, and cryptography~\cite{10.1145/3377929.3389886}. While they provide no guarantee to reach optimal solutions, the obtained solutions are often relevant enough to maintain the claim that EAs are viable approaches for many real-world applications.
Considering the cryptography perspective, common examples of successful applications are evolution of Boolean functions~\cite{10.1145/3449639.3459362}, evolution of S-boxes~\cite{DBLP:journals/ccds/MariotPLJ19}, attacks on PUFs~\cite{10.1007/978-3-662-48324-4_27}, hardware Trojan detection~\cite{DBLP:conf/ches/SahaCNAM15}, and side-channel attacks~\cite{https://doi.org/10.1002/sec.1308}.
From those applications, the evolution of Boolean functions is one of the most successful examples. Indeed, various EAs (as discussed in Section~\ref{sec:related}) have managed to evolve Boolean functions fulfilling diverse cryptographic conditions.
Nevertheless, we can recognize one major problem with those works: they evolve specific Boolean functions, but such an approach often does not generalize to larger Boolean functions with a larger size. To circumvent this problem, one needs to use algebraic constructions that generalize for multiple sizes. There are two types of constructions: primary, where functions are created from scratch, and secondary, where previously constructed functions are used as building blocks~\cite{carlet_2021}.

For a Boolean function to be useful in cryptography, it should be large enough (e.g., minimum 13 inputs), be balanced, and have the highest possible nonlinearity. While these conditions seem reasonable, there are no known algebraic constructions to reach larger nonlinearity than what is obtained with simple quadratic functions (while the existence of sporadic functions shows that it is possible to do so). The reasons for this can be multiple. The most obvious reason is that it is not even known what is the best possible nonlinearity for balanced Boolean functions with more than seven inputs. Also, designing a construction is, in general, a difficult task. As one does not know what kinds of constructions are possible, the search space is prohibitively large. 

Thus, finding such algebraic constructions would be very useful. Besides being a source of Boolean functions for direct applications in cryptography, any developments would also be highly relevant for the research in Boolean functions and error-correcting codes.

The literature is abundant with examples for primary or secondary constructions of bent Boolean functions. Bent functions have the best possible nonlinearity, but they exist only when the number of variables is even, and moreover they are unbalanced. Some of the existing constructions for bent functions have been adapted to generate balanced functions with good nonlinearity (see e.g.~\cite{carlet_2021} for an overview).
EAs also showed capable of evolving constructions of bent Boolean functions~\cite{PicekJ16}. Still, evolving bent Boolean functions seem to be simpler and less practically relevant as bent functions do not have a direct usage in cryptography.

This paper aims to investigate whether genetic programming (GP) can evolve algebraic constructions that, in turn, provide balanced, highly nonlinear Boolean functions. To the best of our knowledge, this is the first work that addresses such an optimization problem using EAs. In our approach, a candidate construction in the GP population is encoded by a tree whose leaves are either seed functions of $n$ variables with high nonlinearity or additional independent variables. The output of the tree is a $(n+k)-$variable Boolean function (with $k$ being the number of additional variables), which is, in turn, evaluated for its balancedness and nonlinearity. We experimentally test our approach using different sizes for the seed functions, using either 1 or 2 additional variables. Our main findings are as follows:
\begin{compactenum}
\item GP can evolve many optimal constructions that achieve the target nonlinearity for relatively small sizes, with the addition of two variables generally performing better than adding a single one.
\item In the experiments where GP obtains optimal constructions with full success rate, many solutions turn out to be the same after minimizing the corresponding circuits and checking for pairwise equivalence.
\item One of the optimal solutions that occur in all considered experiments, with more or less bloated variants, is a particular case of the well-known indirect sum construction~\cite{carlet_2021}.
\end{compactenum}
We finally provide a possible explanation for the third finding, which relates to the way the additional variables are used in the constructions. This prompts us with interesting directions for future research on this optimization problem, which we overview in the conclusions of the paper.

\section{Background}
\label{sec:background}

\subsection{Notation}

Let $n$ be a positive integer, i.e.,  $n \in \mathbb{N}^+$.
We will denote the set of all $n$-tuples of elements in the finite field $\mathbb{F}_{2} = \{0,1\}$ as $\mathbb{F}_{2}^{n}$. We denote the inner product of two vectors $a$ and $b$ by $a \cdot b$, and it equals $a\cdot b = \bigoplus_{i=0}^{n-1} a_{i}b_{i}$. Here, ``$\oplus$'' denotes the addition modulo two (bitwise XOR). The support ($supp$) of a Boolean function $f$ is the set containing the non-zero positions in the truth table representation, i.e., $supp(f) = \left\lbrace x : f(x) = 1\right\rbrace$. The Hamming weight $w_H(f)$ of a Boolean function $f$ equals the size of its support.

\subsection{Boolean Functions}

A Boolean function of $n$ variables is a mapping \textit{f} from $\mathbb{F}_{2}^{n}$ to $\mathbb{F}_{2}$, and it can be uniquely represented by a truth table.
The truth table of $f$ is the vector $(f(0,\cdots,0),\ldots, f(1,\cdots,1))$ containing the function values of $f$, with the input vectors ordered lexicographically.

The Walsh-Hadamard transform $W_{f}$ is another unique representation of a Boolean function. It measures the correlation between $f(x)$ and the linear functions $a\cdot x$~\cite{carlet_2021}:
\begin{equation}
W_{f} (a) = \sum\limits_{x \in \mathbb{F}_{2}^{n}} (-1)^{f(x) \oplus a\cdot x}.
\end{equation}

A Boolean function $f$ is balanced if its truth table vector has the same number of 0s and 1s, or equivalently if $|supp(f)| = 2^{n-1}$.

The minimum Hamming distance between a Boolean function $f$ and all affine functions $a \cdot x \oplus b$ is called the nonlinearity of $f$. The nonlinearity $Nl_{f}$ of a Boolean function $f$ can be expressed in terms of the Walsh-Hadamard coefficients as~\cite{carlet_2021}:
\begin{equation}
\label{eq:nonlinearity}
Nl_{f} = 2^{n - 1} - \frac{1}{2}\max_{a \in \mathbb{F}_{2}^{n}} |W_{f}(a)|.
\end{equation}

The nonlinearity of a Boolean function with $n$ inputs is bounded above by the following inequality:
\begin{equation}
\label{eq:covering_radius_bound}
    Nl_f \leq 2^{n-1}- 2^{\frac{n}{2}-1}.
\end{equation} 
This bound is usually called the Covering Radius Bound.
Note that the so-called \emph{bent} functions satisfy with equality this bound, and therefore they are maximally nonlinear. However, bent functions cannot be balanced and exist only for even $n$, limiting their applicability in cryptography.
When $n$ is odd, the bound given in Eq.~\eqref{eq:covering_radius_bound} cannot be tight and the maximal nonlinearity lies between $2^{n-1} - 2^{\frac{n-1}{2}}$ and $2^{n-1}- 2^{\frac{n}{2}-1}$. Here, $2^{n-1}-2^{\frac{n-1}{2}}$ is also called the quadratic bound because it is the best nonlinearity achievable by quadratic functions (i.e. functions of algebraic degree $2$). The maximum possible nonlinearity for balanced Boolean functions is unknown for all $n>7$. Table~\ref{tab:maxNL} recaps the optimal and best-known nonlinearities values for such functions of several sizes.

\begin{table*}[]
\caption{Best known nonlinearity values for balanced Boolean functions. Entries in bold are optimal.}
\label{tab:maxNL}
\adjustbox{max width=\columnwidth}{%
\begin{tabular}{@{}lccccccccccccc@{}}
\toprule
Variables & 4 & 5  & 6  & 7  & 8   & 9   & 10  & 11  & 12   & 13   & 14   & 15    & 16    \\ \midrule
Max NL    & {\bfseries 4} & {\bfseries 12} & {\bfseries 26} & {\bfseries 56} & 116 & 240 & 492 & 992 & 2012 & 4036 & 8120 & 16272 & 32638 \\ \bottomrule
\end{tabular}
}
\end{table*}

If a Boolean function is not balanced, it cannot be used in cryptography as it causes a statistical bias. Similarly, if a Boolean function is not highly nonlinear, it will not provide optimal (or near to optimal) resilience against linear cryptanalysis.

\subsection{Construction Techniques}
\label{subsec:construction}

There are three viable options to create Boolean functions: algebraic constructions, random search, and heuristic approaches.
The main strength of algebraic constructions is that they generate functions with certain properties, and it is equally easy to construct functions of any number of variables. 
The main drawback lies in the fact that they are deterministic and always result in the same functions up to affine equivalence (that is, up to the composition by an affine automorphism), which means the number of different functions one can obtain is limited. 
Furthermore, it is quite difficult to devise an algebraic construction that results in Boolean functions with the desired properties.
Heuristic methods are known to generate a large number of good results in a relatively short time. However, the search space size grows exponentially with the number of variables, and it becomes difficult to work even with a moderate number of inputs.

The construction techniques can be divided into \emph{primary} constructions and \emph{secondary} constructions.
In primary constructions, one obtains new functions without using known ones. In secondary constructions, one uses existing functions to construct new ones~\cite{carlet_2021}. 

The Rothaus construction represents an example of a secondary algebraic construction~\cite{Carlet_bent}.
Let $h_1, h_2,$ and $h_3$ be three bent functions with $n$ inputs, with $h_1 \oplus h_2 \oplus h_3$ also being a bent function. A new bent function of $n + 2$ variables is generated as:
\begin{eqnarray}
f(x, x_{n+1}, x_{n+2}) = h_1(x)h_2(x) \oplus h_1(x)h_3(x) \\ \nonumber
\oplus h_2(x)h_3(x) \oplus [h_1(x) \oplus h_2(x)]x_{n+1} \\ \nonumber
\oplus [h_1(x) \oplus h_3(x)]x_{n+2} \oplus x_{n+1}x_{n+2}.
\end{eqnarray}

This kind of construction is a motivation for our approach, where we aim to evolve secondary algebraic constructions that use a number of $n$-variable highly nonlinear balanced Boolean functions to produce either $(n + 1)$ or $(n + 2)$-input balanced functions having high nonlinearity.

\section{Related Works}
\label{sec:related}

The history of using evolutionary algorithms to evolve Boolean functions with good cryptographic properties is already 25 years long. As far as we know, the first work using EAs for Boolean functions with specific cryptographic properties happened in 1997. There, the authors used genetic algorithms to evolve Boolean functions with high nonlinearity~\cite{Millan97}.
Next, various algorithms were tested to obtain even better results. For instance, Millan et al. used GA in combination with hill climbing and a resetting step to evolve highly nonlinear Boolean functions up to 12 inputs~\cite{millan}. 
On the other hand, Clark and Jacob used simulated annealing and hill-climbing with a cost function motivated by the Parseval theorem to find  Boolean functions with high nonlinearity and low autocorrelation~\cite{clark_two_stage}.
Aguirre et al. were the first to consider multi-objective optimization for this problem~\cite{hernan}. The authors used a random bit climber to find balanced, highly nonlinear Boolean functions.

Picek et al. considered various EAs (genetic algorithms and genetic programming) to find Boolean functions that fulfill multiple cryptographic properties~\cite{picekgecco2013}. 
Mariot and Leporati experimented with Particle Swarm Optimization~\cite{Kennedy1995} to find Boolean functions with good trade-offs of cryptographic properties for sizes up to 12 inputs~\cite{MariotL15}. 
Picek et al. investigated various immunological algorithms to evolve highly nonlinear Boolean functions up to 16 inputs~\cite{PicekSJ17}. Clark et al.~\cite{clark04} pioneered the spectral inversion approach where pseudo-Boolean functions are represented by Walsh spectra that satisfy good cryptographic properties; the optimization objective is to find a spectrum that corresponds to a true Boolean function. Mariot and Leporati~\cite{mariot15} further investigated this approach by proposing a genetic algorithm to evolve such spectra.

The above-listed works make only a small part of the research done but show how most of the works manage to find highly fit Boolean functions (whatever the properties required). Still, there was always the problem of using such computationally heavy approaches for Boolean functions with more inputs or cryptographic properties that are more expensive to evaluate. Additionally, such approaches resulted in specific Boolean function instances, running searches for every new size required.

Picek and Jakobovic considered an approach where instead of evolving Boolean functions, they evolved constructions resulting in Boolean function with the required properties~\cite{PicekJ16}. They used genetic programming to evolve secondary algebraic constructions of bent (thus, imbalanced but maximally nonlinear) Boolean functions~\cite{PicekJ16}.
Carlet et al. used genetic programming to improve Boolean functions obtained through algebraic constructions~\cite{10.1145/3449639.3459362}. This approach resulted in Boolean functions obtained through the Hidden Weight Boolean Function construction with higher nonlinearity than previously known. Finally, Mariot et al.~\cite{mariot21} investigated a secondary construction based on cellular automata (CA), using evolutionary strategies to search for CA local rules that result in bent and semi-bent functions when plugged into the construction.

For a more detailed overview of EAs and Boolean functions in cryptography, we refer interested readers to~\cite{10.1145/3377929.3389886}.

\section{Methodology}
\label{sec:methodology}

This section describes how to evolve Boolean functions from scratch using GP and then evolve secondary constructions that rely on predefined Boolean functions in a smaller size.

\subsection{Evolving Boolean Functions with GP}

GP and its variants (most notably Cartesian Genetic Programming~\cite{10.5555/2934046.2934074}) have already been extensively used in the evolution of Boolean functions as indicated in Sec.~\ref{sec:related} and have been able to produce human-competitive results.
As a baseline approach, we use GP to evolve a function in the symbolic form, using a tree representation.
According to the truth table it produces, each tree is evaluated for the nonlinearity property.
The terminal set is comprised of a given number of Boolean variables, which we denote with $v_0, v_1, ..., v_{n-1}$.
The function set consists of several Boolean primitives, which can be used to represent any Boolean function.
Our experiments use the following function set: OR, XOR, AND, AND2, XNOR, and function NOT that takes a single argument.
The function AND2 behaves the same as the function AND but with the second input inverted.
Additionally, we included the function IF, which takes three arguments and returns the second one if the first one evaluates to true and the third one otherwise.

\subsection{Evolving Boolean Constructions}
\label{subsec:constructions}

\begin{figure}
\centering
\includegraphics[scale=0.75]{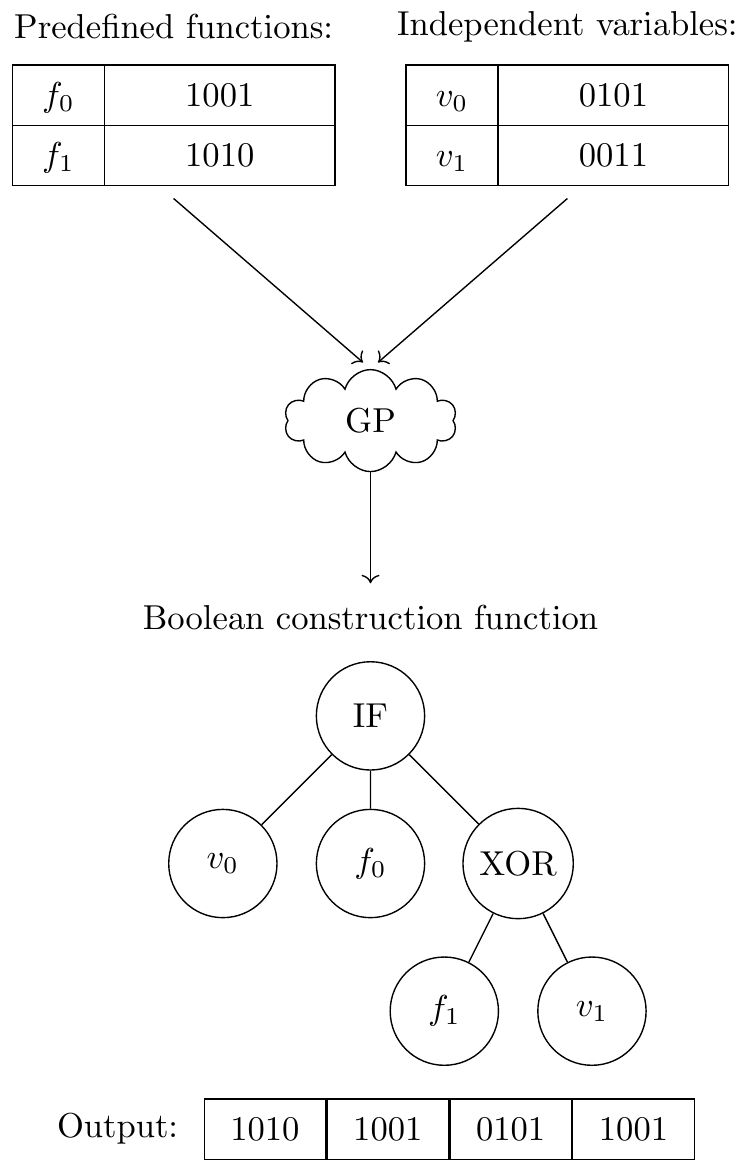}
\caption{Outline of evolving Boolean constructions ($2$ seed functions of $2$ variables, construction resulting in $n + 2 = 4$ variable Boolean function)}
\label{fig:diagram}
\end{figure}

To evolve constructions with GP, we take a slightly different approach.
Firstly, we presume the existence of a certain number of predefined Boolean functions (seed functions) that are included in the terminal set.
In our experiments, up to four predefined Boolean functions are available as terminals, which are denoted with $f_0$, $f_1$, $f_2$, and $f_3$.
The number of variables of seed functions is taken to be $n$, and they are given by their truth tables.
Additionally, the terminal set includes a number of independent Boolean variables; if a single variable is added ($v_0$), then the resulting construction (a GP tree) represents a new Boolean function with $n + 1$ variables.
Likewise, with two Boolean variables, $v_0$ and $v_1$, the construction obtains an $(n + 2)$-variable function.
The function set remains the same as in the previous method. Figure \ref{fig:diagram} shows the outline of the entire construction process, in which the GP based on a set of predefined functions and input variables constructs a Boolean function. 

To be able to apply this approach, the seed functions must be given or previously evolved with the general GP method.
Since we optimize for high nonlinearity, the seed functions are presumed to be optimally (where possible) or highly nonlinear.
The initial set of seed functions is obtained with the general method, starting with a low number of variables (e.g., four variables), which is trivial to find.
Then, the seed functions are used to find constructions for a larger number of variables (e.g., six).
The evolved constructions can be decoded and stored as a truth table; that way, the outputs of the previous stage may then be used as seed functions in the next stage, in a kind of bootstrap procedure.

\subsection{Fitness Functions}
\label{subsec:fitness}

The algorithm should find constructions that give balanced, highly nonlinear functions. 
To facilitate this, we distinguish the \textit{objective value} of the resulting Boolean function and the \textit{fitness value of the construction} that obtained this function.

We use a two-stage objective function in which a bonus equal to the nonlinearity is awarded only to a perfectly balanced function; otherwise, the objective value is only described by the balancedness penalty. 
The balancedness penalty $BAL$ is defined as the difference up to the balancedness (i.e., the number of bits to be changed to make the function balanced). 
This difference is included in the objective function with a negative sign to act as a penalty in maximization scenarios.
The delta function $\delta_{BAL, 0}$ assumes the value one when $BAL = 0$ and is zero otherwise. 
\begin{equation}
\label{eq:obj1}
objective_1 : -BAL + \delta_{BAL, 0} \cdot Nl_{f}.
\end{equation}

The second objective function extends the first one to consider the whole Walsh-Hadamard spectrum:
\begin{equation}
\label{eq:obj2}
objective_2 : -BAL + \delta_{BAL, 0} \cdot (Nl_{f} + Indicator).
\end{equation}
In this expression, the $Indicator$ term is the normalized number of occurrences of the largest nonlinearity value in the whole spectrum (denoted $\#max\_values$). The smaller the number of these largest values, the easier it is for the algorithm to reach the next nonlinearity value:
$
Indicator = 1 - \frac{\#max\_values}{2^n}.
$

When evolving constructions, we aim to obtain a general construction that will be able to produce a highly nonlinear function for every \textit{combination} of seed functions of lower order.
For this purpose, we evaluate the constructions with several \textit{groups} of seed functions, where each group consists of different values of terminals $f_0 - f_3$.
In our experiments, we use four groups of seed functions, where for each group $i$, the value of the objective function is calculated with the same tree (i.e., the same construction).
The resulting fitness for the evaluated construction is then defined in three ways.
\begin{enumerate}
\item[A)] The first method considers the objective value obtained with the first seed group; only if the nonlinearity reaches a predefined level (e.g., the best-known value), then the other seed groups are used, and their obtained objective value is added to the first one to obtain the fitness value:
\begin{equation}
\label{eq:fit1}
fit_1 : val_1 + \delta_{val_1, targetVal} \cdot \sum val_i.
\end{equation}

\item[B)] The second approach sums the objective value obtained by all the seed groups; we denote this as \textit{sum of all groups}: 
\begin{equation}
\label{eq:fit2}
fit_2 : \sum val_i.
\end{equation}

\item[C)] The third method considers the \textit{minimum} objective value among all seed groups, which is maximized as a consequence:
\begin{equation}
\label{eq:fit3}
fit_3 : \min val_i.
\end{equation}

\end{enumerate}

Naturally, this approach does not guarantee that the evolved construction will be general; thus, every evolved construction is subsequently evaluated with a separate test set of seed functions.

Finally, we observe that when evolving constructions, the obtained trees with maximal fitness always include the two Boolean variables $v_0$ and $v_1$, but not necessarily the whole set of input functions $f_0 - f_3$.
To find meaningful expressions that can be candidates for general construction (see Sec.~\ref{subsec:construction}), we need to ensure that all input seed functions are contained in every construction. Therefore, we add a penalty step, in which a construction is penalized if it does not include all the input terminals.
This penalization is applied to all the fitness functions and can be represented with the following equation:
\begin{equation}
	\label{eq:penalization}
	fitness_i =  \frac {fitness_i} {1 + missing\_terminals},
\end{equation}
which simply equals to $fitness_i$ divided by the number of missing input terminals.

\begin{table}[]
\centering
\caption{}
\label{tab:variants}
\adjustbox{max width=\columnwidth}{%
\begin{tabular}{@{}p{0.5\linewidth}p{0.5\linewidth}@{}}
\toprule
Parameter description          & Parameter value \\ \midrule
Number of variables of target Boolean function & 5, 6, 7, 8                                                     \\
Independent variables          & 1, 2            \\
Number of seed functions       & 2, 4            \\
Number of seed function groups & 4               \\
Seed functions type            & balanced, bent  \\
Objective value                & nonlinearity (\ref{eq:obj1}), nonlinearity with spectrum (\ref{eq:obj2})  \\
Type of fitness function       & first group (\ref{eq:fit1}), sum of all groups (\ref{eq:fit2}), minimum of all groups (\ref{eq:fit3})\\
\bottomrule
\end{tabular}%
}
\end{table}

\subsection{Experimental Settings}
\label{sec:settings}

The parameters for the GP are the same for all configurations and are based on our previous experience, as well as guidelines from the existing literature addressing similar problems.
The population size is set to 500, and the maximal tree depth to 5.
We employ a steady-state selection operator with a 3-tournament elimination, which in each iteration randomly selects three individuals for the tournament and eliminates the worst one. 
A new individual is created immediately by crossing over the remaining two from the tournament, which then undergoes mutation with a probability of 0.5. 
The variation operators used for GP are simple tree crossover, uniform crossover, size fair, one-point, and context preserving crossover~\cite{poli08:fieldguide} (selected at random) and subtree mutation. 

Common parameters for all the experiments include the termination condition of 500\,000 fitness evaluations. We chose this particular bound because our preliminary tests showed that final solutions are mostly found before reaching this number of evaluations. Finally, each experiment is repeated 30 times.

\section{Experimental Results}
\label{sec:results}

In this section, we present the results of the described experiments.
First, we applied a canonical GP to evolve balanced, highly nonlinear Boolean functions.
We limit to functions with a larger number of variables; for $n < 8$, the optimal nonlinearity values are known (see Table~\ref{tab:maxNL}), and GP has no difficulties in finding functions with this property.
For this experiment only, the number of runs was set to 100, and the results are shown in Table~\ref{tab:GP}.

We can see that, for some function sizes, the search always converges to the same level of nonlinearity (i.e., for $n = 9, 11, 17$).
In most cases, however, the GP managed to obtain the best-found value only with a lower rate (i.e., one or two out of 100 runs).

For the constructions, the experiments were performed in two phases. In the first phase, we conducted experiments in which we explored all the configuration variants presented in Table~\ref{tab:variants}. 
For example, to get to construction that results with 8 variables, we used seed functions of both 6 variables (adding two independent ones, $n+2$) and 7 variables (plus a single independent one, $n+1$); the number of seed functions was 2 or 4, and they were either balanced or bent; objective value used two options, and all three fitness functions were tested.
In cases where balanced highly nonlinear seed functions were used, their nonlinearity was equal to that from Table~\ref{tab:maxNL}, since they are relatively easy to obtain with GP.
Also, in all cases when using the first fitness function, best known nonlinearity value from Table~\ref{tab:maxNL} is used as a $targetVal$ in~\ref{eq:fit3}.

The first phase aimed to identify configurations that allow us to obtain ``locally optimal'' constructions; we define those as the ones that manage to reach the target nonlinearity value for a given number of variables (5-8). The first thing to note is that there is no comparative advantage in using the objective value~(\ref{eq:obj1}), which only considers nonlinearity; all the experiments show that~(\ref{eq:obj2}) is never worse and almost always a better choice. Secondly, it seems that constructions of the form $n+1$ are generally worse than the ones introducing two variables ($n+2$), as locally optimal solutions are seldomly found in the former. Using bent seed functions produces the same results as the balanced seeds for target sizes 5 and 6 but obtains worse results in sizes 7 and 8. Finally, when only 2 input seed functions are used, the first fitness function does not find locally optimal construction; otherwise (for 4 seed functions), there is no difference between them. It is interesting that no construction has produced functions with nonlinearity 26 (which is optimal) for 6 variables. This is surprising since those functions can be found relatively easily with a search-based approach.

In the second phase, we take all the configurations that were able to produce locally optimal constructions and evaluate them on a separate test set of seed functions.
The test set comprises of 8 groups of balanced highly nonlinear functions, obtained either with search-based GP (up to 13 variables) or with the constructions themselves, using a bootstrap approach.
In this phase, the \textit{same} construction (taken from a single run) is used to produce the resulting Boolean function whose nonlinearity is then evaluated on the whole range of variables from $n=6$ to $n=18$.

In this experiment, it became evident that the $n+1$ constructions we found are not general since they do not reach target values where the number of variables is different from the one on which they were evolved; the same can also be observed for constructions using bent seeds.
All the other configurations (i.e., with two independent variables ($n+2$), using balanced seeds) could produce ``general'' constructions in at least several runs.

The results for those constructions are presented in Table~\ref{tab:constructions}; the first row lists the target number of variables ($n+2$), the middle row shows the nonlinearity of seed functions in $n$ variables, and the last row shows the obtained nonlinearity.
The most important finding in this phase is that, although the evolved constructions look different (with a different genotype) and produce different resulting Boolean functions, all the constructions we tested always produce the same nonlinearity for a given number of variables.
We try to analyze this behavior in the next section.

From the results, it is evident that the obtained constructions are ``general'', in the sense that they always produce balanced Boolean functions with very high nonlinearity.
In some cases, that nonlinearity is optimal (which is known for $n<8$) and in some cases equal to the one obtained by search-based GP.
It is interesting to note that, in cases where constructions produce lower nonlinearity than simple GP, the GP has a significantly lower probability of finding better solutions.

\begin{table*}[]
\centering
\caption{Results for search-based GP in finding balanced highly nonlinear Boolean functions}
\label{tab:GP}
\adjustbox{max width=\columnwidth}{%
\begin{tabular}{@{}lllllllllll@{}}
\toprule
GP search         & n = 9 & n = 10 & n = 11 & n = 12  & n = 13  & n = 14 & n = 15  & n = 16  & n = 17 & n = 18 \\ \midrule
min NL            & 240   & 480    & 992    & 1\,953    & 3\,905    & 8\,001   & 16\,192   & 32\,512   & 65\,280  & 130\,561 \\
avg NL            & 240   & 484.24 & 992    & 1\,996.69 & 4\,028.39 & 8\,087.3 & 16\,253.5 & 32\,542.2 & 65\,280  & 130\,622 \\
max NL            & 240   & 492    & 992    & 2\,008    & 4\,032    & 8\,120   & 16\,256   & 32\,608   & 65\,280  & 130\,753 \\
\#max & 100\% & 2\%    & 100\%  & 14\%    & 96\%    & 1\%    & 97\%    & 2\%     & 100\%    & 2\%    \\ \bottomrule
\end{tabular}
}
\end{table*}

\begin{table*}[]
\centering
\caption{Results for secondary constructions producing balanced highly nonlinear Boolean functions}
\label{tab:constructions}
\adjustbox{max width=\columnwidth}{%
\begin{tabular}{@{}lllllllllllll@{}}
\toprule
constructions & n = 7 & n = 8 & n = 9 & n = 10 & n = 11 & n = 12 & n = 13 & n = 14 & n = 15 & n = 16 & n = 17 & n = 18 \\ \midrule
seed NL       & 12    & 26    & 56    & 116    & 240    & 488    & 992    & 2\,000   & 4\,032   & 8\,096   & 16\,256  & 32\,576  \\
resulting NL  & 56    & 116   & 240   & 488    & 992    & 2\,000   & 4\,032   & 8\,096   & 16\,256  & 32\,576  & 65\,280  & 130\,688 \\ \bottomrule
\end{tabular}
}
\end{table*}

\section{Discussion}
\label{sec:discussion}

We now investigate in detail the constructions produced by GP for those experiments that always converged to a general solution in each run. In particular, we considered five experiments with two seed functions and three experiments with four seed functions. In all considered experiments, the constructions extended the seed functions by two additional variables. In what follows, an experiment is synthetically identified by the tuple $(s, n, nl, ev)$, where $s$ is the number of seed functions used as input for the constructions, $n$ is the number of variables of each seed function (which means that the construction will generate functions of $n+2$ variables), $nl$ is the nonlinearity of the seed functions, and $ev$ is the evaluation method used by GP to evolve the constructions. In particular, we denote by A, B, C respectively the three fitness functions $fit_1$, $fit_2$ and $fit_3$ defined in Section~\ref{subsec:fitness}.

\subsection{Solutions Size}
\label{subsec:sol_size}

\begin{figure}
\centering
\includegraphics[width=1.1\columnwidth]{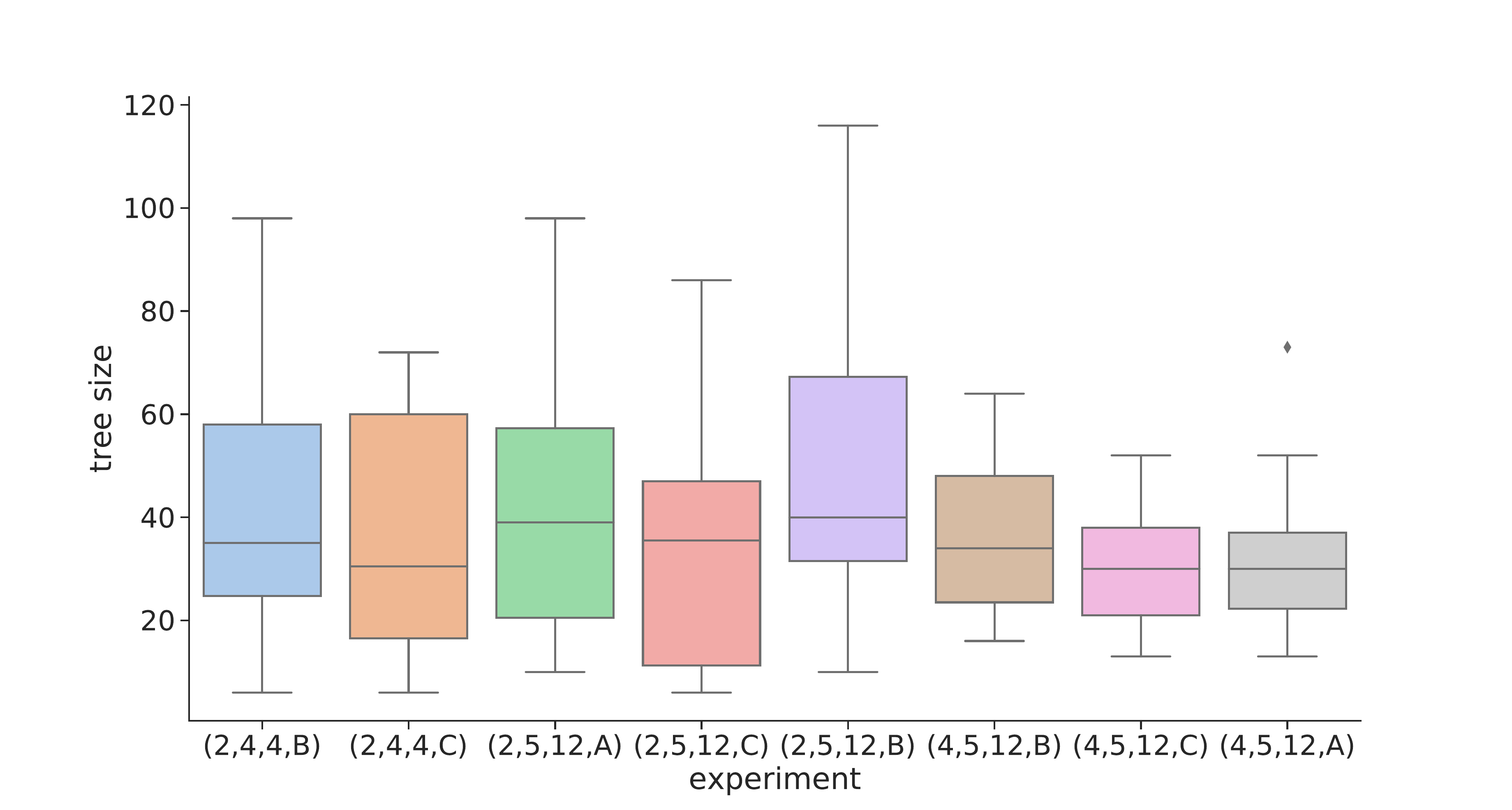}
\caption{Tree size distribution for the eight experiments that always converged to a general solution.}
\label{fig:boxplot_size}
\end{figure}

We start by analyzing the size of the trees generated by GP, considering it as an interpretability proxy. In particular, we define the size of a tree as the number of its nodes, independently of the type of functional or input variable (i.e., additional independent variable or seed function). Figure~\ref{fig:boxplot_size} plots the distributions of the tree sizes for all eight experiments.

The first interesting remark is that the distributions of the five experiments with two seed functions are quite dispersed, with extreme values ranging as low as 6 nodes and as high as 120 nodes. Further, the upper quartiles of these distributions make up a large part of the interquartile ranges, except for the experiment $(2,5,12,C)$. Considering also that the median tree size is around 35-40 nodes, we can conclude that most of the trees evolved by GP with two seed functions are too unwieldy to be interpreted by hand. The situation seems better with the three experiments using four seed functions, where the upper and lower quartiles are more balanced, and there is also a smaller difference between the minimum and maximum values. However, the median tree sizes are similar to those of the experiments with two seed functions, and the minimum sizes are significantly higher. Therefore, we end up with constructions that are quite difficult to interpret also with four seed functions.

\subsection{Solutions Diversity}
\label{subsec:sol_div}

As a next step, we employed the ESPRESSO heuristic logic minimizer~\cite{Rudell87} to simplify the GP trees obtained in all eight experiments, with a twofold objective. First, we determined the simplest possible circuit of each construction by performing an exact minimization and verified if its size was small enough to elicit a manual interpretation. Unfortunately, the resulting minimized expressions were still too complex for a deeper analysis. As a second objective, we checked for \emph{equivalent} circuits to investigate how many different solutions GP can generate within each experiment. In particular, the ESPRESSO tool allows checking if two different circuits are equivalent by comparing their truth tables and applying basic equivalence relations such as output negation or permutation of the input variables. We performed a pairwise equivalence test among all solutions evolved by GP in each considered experiment and built the corresponding graphs. Hence, each graph is composed of 30 nodes (one node per solution), and two nodes are connected by an undirected edge if and only if the two circuits were marked as equivalent by the ESPRESSO minimizer.

\begin{figure}[t]
\centering
\includegraphics[width=\columnwidth]{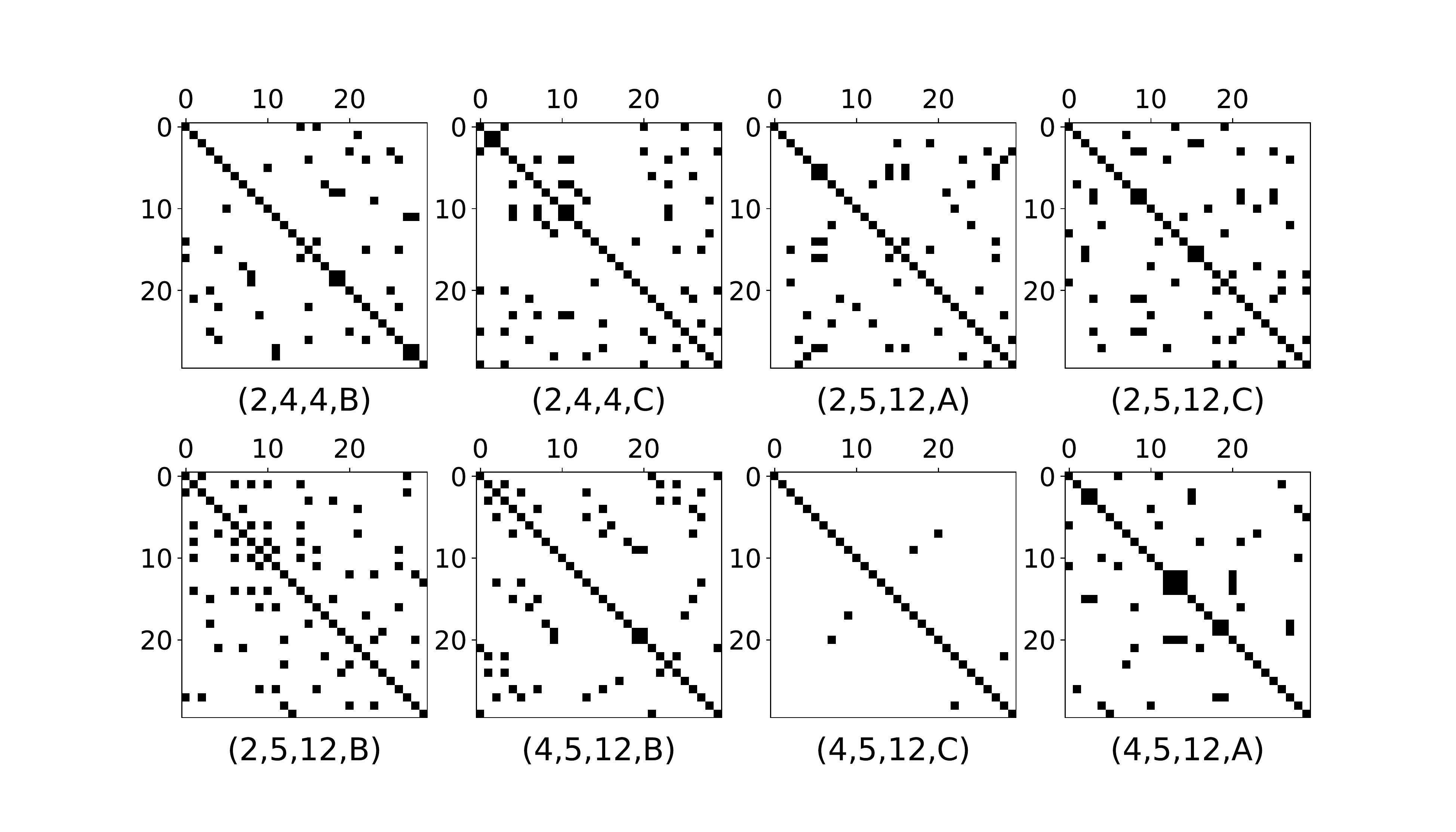}
\caption{Adjacency matrices for the equivalence relation graphs of all eight experiments.}
\label{fig:adjacency}
\end{figure}

Figure~\ref{fig:adjacency} displays the adjacency matrices of the equivalence relation graphs, while Table~\ref{tab:equiv} reports for each experiment the number of distinct solutions (or equivalence classes), the size of the largest equivalence class, and the number of seeds effectively used.

\begin{table}[t]
\centering
\caption{Equivalence classes summary for all experiments. Bold values represent experiments where all constructions use \emph{less} seeds than allowed.}
\label{tab:equiv}
\begin{tabular}{cccc}
\toprule
Exp. & \#classes & max\_size & seeds\_used \\
\midrule
(2,4,4,B) & 15 & 4 & 2 \\
(2,4,4,C) & 13 & 5 & 2 \\
(2,5,12,A) & 15 & 5 & 2 \\
(2,5,12,C) & 13 & 5 & 2 \\
(2,5,12,B) & 11 & 5 & 2 \\
\midrule
(4,5,12,B) & 14 & 4 & {\bfseries 2} \\
(4,5,12,C) & 27 & 2 & 4 \\
(4,5,12,A) & 14 & 4 & {\bfseries 2} \\
\bottomrule
\end{tabular}
\end{table}
In general, one can see from Figure~\ref{fig:adjacency} and Table~\ref{tab:equiv} that using two seed functions generally leads GP to evolve many equivalent functions, with $(2,5,12,B)$ yielding the smallest number of distinct solutions and the largest equivalence classes. This is somewhat expected since the number of ways to combine the two additional variables of the constructions is smaller with two seed functions than with four. Accordingly, the experiment $(4,5,12,C)$ is the one giving the highest diversity among solutions, with only 6 equivalent solutions grouped in three equivalences classes of size 2.

Contrarily, the distributions of equivalence classes for the other two experiments with four seed functions, namely $(4,5,12,B)$ and $(4,5,12,A)$, are closer to those with two seeds. What is more surprising is that all solutions evolved by GP in these two experiments actually use \emph{less} seeds than expected, as shown by the entries in bold of Table~\ref{tab:equiv}. Although the original GP trees of each solution in these two experiments use all seeds functions, the ESPRESSO tool always returned minimized circuits where the seeds $f_2$ and $f_3$ are never used. This is interesting since, as explained in Section~\ref{subsec:fitness}, we adopted a penalty factor in all our fitness functions in order to force the occurrence of all seed functions in the candidate trees. However, in this particular experiment GP was able to circumvent this penalty by using all four seed terminals at a syntactic level, but encoding two of them in subtrees that do not affect the output of the constructions. The cause of this phenomenon likely resides in the underlying fitness functions, which are $fit_2$ and $fit_1$ respectively for the experiments $(4,5,12,B)$ and $(4,5,12,A)$. In both cases, each seed group can contribute in a non-uniform way to the fitness value of an individual. This might, in turn, lead the GP evolutionary process to favor general constructions with fewer active seed functions. On the other hand, the experiment $(4,5,12,C)$, where all four seed functions partake in the minimized circuits, is based on the fitness function $fit_3$, which maximizes the minimum objective value among all seed groups. In this case, GP is forced to evolve constructions that yield highly nonlinear balanced functions for each tested group, possibly increasing the chances that all seed functions are combined uniformly.

\subsection{Interpreting Simple Constructions}
\label{subsec:interp}

So far, we discussed the general constructions concerning their sizes and diversity, which gave us some insights on the GP's behavior for this particular optimization problem. We now investigate the specific nature of these constructions to determine if they are new or already known in the literature of Boolean functions.

As remarked in Section~\ref{subsec:sol_div}, the minimized circuits obtained through ESPRESSO are, on average, still too complex to allow a manual interpretation. For this reason, here we analyze in detail only some of the simplest constructions evolved by GP. In particular, we selected one construction for each of the eight experiments investigated in the previous sections. Our selection criteria for "simplicity" were as follows:
\begin{compactenum}
\item Small tree size, by considering the lower quartiles of the tree size distributions in Figure~\ref{fig:boxplot_size} as an upper bound.
\item IF node at the root, so the construction is piecewise-defined.
\item Condition at the root IF composed of a single literal (independent additional variable or seed function). This helps avoiding bloated expressions that control the output of the functions resulting from the construction.
\end{compactenum}
Table~\ref{tab:constr} reports the selected constructions for each experiment, as evolved by GP. The notation used for the expressions includes $v_0,v_1$ for the two additional variables and $f_0-f_3$ for the seed functions.
\begin{table}[t]
\centering
\caption{Simplest GP constructions selected for analysis.}
\label{tab:constr}
\begin{tabular}{ccl}
\toprule
exp. & size & construction \\
\midrule
(2,4,4,B) & 6 & IF($v_0$, $f_0$, ($v_1$ XOR $f_1$))\\
(2,4,4,C) & 6 & IF($v_0$, $f_0$, ($f_1$ XOR $v_1$)) \\
(2,5,12,A) & 10 & IF($v_0$, $f_1$, (($v_1$ XOR $f_0$) OR ($v_1$ AND $v_0$))) \\
(2,5,12,C) & 6 & IF($v_1$, $f_1$, ($f_0$ XOR $v_0$)) \\
(2,5,12,B) & 10 & IF(NOT(NOT($v_0$)), NOT(($f_0$ XOR NOT($v_1$))), $f_1$) \\
\midrule
(4,5,12,B) & \multirow{2}{*}{17} & IF($v_1$, ($v_0$ XOR ($f_1$ AND $v_1$)), IF($v_1$, ($f_2$ OR \\
 & & ($f_2$ AND ($f_2$ OR $f_3$))), $f_0$)) \\
(4,5,12,C) & \multirow{2}{*}{17} & IF($v_0$, ($f_1$ XOR $v_1$), (((($f_0$ OR $f_3$) AND2 IF($f_3$, $f_2$,\\  & & $v_1$)) AND $v_0$) OR $f_3$)) \\
(4,5,12,A) & \multirow{3}{*}{22} & IF($v_0$, ($v_0$ AND2 $f_1$), (($v_0$ AND (($f_2$ XOR $v_1$) \\
  & & XOR $f_3$)) XOR (NOT($f_0$) XOR IF(($v_0$ XNOR $v_1$), \\
  & & $v_1$, $v_0$)))) \\
\bottomrule
\end{tabular}
\end{table}
Regarding the constructions with two seed functions, one can easily see that the smallest solutions of size 6 all correspond to the same construction, up to a swap of the XOR operands or a renaming of the leaf nodes. Figure~\ref{fig:tree-cons} depicts the tree of this construction and the corresponding mathematical definition, taking as a reference the solution of the experiment $(2,4,4,B)$.
\begin{figure}[t]
    \centering
    \begin{minipage}{.4\columnwidth}
    \begin{tikzpicture}[scale=0.9]
    \tikzset{every tree node/.style={draw, thick, circle,inner sep=1pt, minimum size=0.7cm}}
    \tikzset{every leaf node/.append style={fill=black!10}}
    \tikzset{edge from parent/.append style={thick}}
        \Tree [.IF $v_0$ $f_0$ [ .$+$ $v_1$ $f_1$ ]
        ]
    \end{tikzpicture}
    \end{minipage}%
    \begin{minipage}{.6\columnwidth}
    \begin{equation}
    \label{eq:conc-cons}
    F(v_0,v_1,v) = 
    \begin{cases}
        f_0(v) \enspace , & \textrm{ if } v_0 = 1 \enspace , \\
        f_1(v) \oplus v_1 \enspace, & \textrm{ if } v_0 = 0 \enspace .
    \end{cases}
    \end{equation}
    \end{minipage}
    \caption{Smallest GP construction. XOR is denoted by $+$.}
    \label{fig:tree-cons}
\end{figure}
Considering the truth table representation, this construction basically concatenates the two seed functions as $f_0||f_0||\overline{f_1}||f_1$, where $\overline{f_1}$ denotes the negation of $f_1$. For this reason, in what follows we refer to this expression as the concatenation construction.

It is interesting to remark that \emph{the remaining five expressions in Table~\ref{tab:constr} still correspond to the concatenation construction}. This can be easily verified for the expressions with two seed functions. In particular, for the experiment $(2,5,12,A)$ the innermost AND always maps to $0$, since we are in the subtree where $v_0=0$, and thus the subsequent OR evaluates to $v_1$ XOR $f_0$. Similarly, for the expression selected in experiment $(2,5,12,B)$, the condition on the root IF is simply a double negation of $v_0$. Further, the subtree NOT($f_0$ XOR NOT($v_1$)) which is selected when $v_0=1$ is equivalent to $f_0$ XOR $v_1$, by comparing the respective truth tables.

\begin{figure}[t]
    \centering
    \begin{tikzpicture}[scale=0.8]
    \tikzset{every tree node/.style={draw, thick, circle,inner sep=1pt, minimum size=0.7cm}}
    \tikzset{every leaf node/.append style={fill=black!10}}
    \tikzset{edge from parent/.append style={thick}}
        \Tree [.IF $v_0$ [ .$\land^2$ $v_0$ $f_1$ ] [ .$+$ [ .$\land$ $v_0$ [.$+$ [ .$+$ $f_2$ $v_1$ ] $f_3$ ] ] [ .$+$ [ .$\lnot$ $f_0$ ] [.IF [ .$\overline{+}$ $v_0$ $v_1$ ] $v_1$ $v_0$ ] ] ]
        ]
    \end{tikzpicture}
    \caption{Example of bloated GP construction. AND2 is denoted by $\land^2$, XNOR by $\overline{+}$.}
    \label{fig:bloated}
\end{figure}
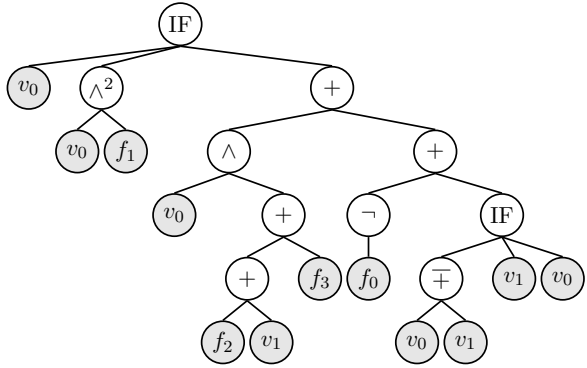

Concerning the three constructions with four seed functions, the simplification process is more convoluted, so we do not report it here in full for the sake of brevity. As an example, we only show in detail the largest tree in Figure~\ref{fig:bloated}, namely the construction selected for $(4,5,12,A)$.
The subtree with AND2 always evaluates to NOT($f_1$), since $v_0=1$ in that branch of the root IF. On the contrary, the AND subtree can be pruned since $v_0=0$, and thus it always evaluates to 0. Hence, the seed functions $f_2$ and $f_3$ are effectively discarded, since they only occur in this prunable subtree. Replacing the other occurrences of $v_0$ with $0$ in the remaining XOR subtree, one finally gets NOT($f_1$) XOR $v_1$, which is a trivial variation of $f_1(v) \oplus v_1$ when $v_0=0$. Therefore, this tree is equivalent to the concatenation construction as well.

Since all solutions analyzed up to now are equivalent, it makes sense to determine whether the concatenation construction corresponds to a known result in the related literature. In particular, taking the expression in Figure~\ref{fig:tree-cons} and exchanging the indices of $v_0$ and $v_1$, one can see that this construction is a particular case of the \emph{indirect sum construction}~\cite{carlet_2021}, where only two additional variables are used to extend the functions. The following result, of which we omit the proof since it is just a trivial adaptation of Proposition 83 in~\cite{carlet_2021}, characterizes the Walsh spectrum and the nonlinearity of the functions resulting from this construction.
\begin{lemma}
\label{lm:constr-gp}
Let $F: \F_2^{n+2} \to \F_2$ be defined as in Eq.~\eqref{eq:conc-cons}. Then, for any $(v_0,v_1,v) \in \F_2^{n+2}$ (with $v_0,v_1 \in \F_2$ and $v \in \F_2^n$) the Walsh coefficient $W_F(v_0,v_1,v)$ is equal to:
\begin{equation}
\label{eq:walsh-constr}
W_F(v_0, v_1, v) =
\begin{cases}
(-1)^{v_0} \cdot 2\cdot W_{f_1}(v) & , \textrm{ if } v_1=0 , \\
2\cdot W_{f_2}(v) & , \textrm{ if } v_1=1 .
\end{cases}
\end{equation}
Further, if $f_1$ and $f_2$ are both balanced and have the same nonlinearity $nl = 2^n - \frac{1}{2} \cdot S$ ($S$ being the maximum Walsh coefficient in absolute value of $f_1$ and $f_2$), then $F$ is also balanced and its nonlinearity equals:
\begin{equation}
\label{eq:nl-cons}
nl(F) = 2^{n+1} - S \enspace .
\end{equation}
\end{lemma}
In particular, Equation~\eqref{eq:nl-cons} gives a theoretical explanation to the fact that the concatenation construction is general, or equivalently that it is optimal for the fitness functions adopted in our GP experiments.

\section{Conclusions and Future Work}
\label{sec:conclusions}

This paper proposed for the first time a GP approach to evolve secondary constructions of Boolean functions that are both balanced and highly nonlinear, which are particularly relevant in the design of symmetric ciphers. A candidate construction is encoded by a tree where the internal nodes are Boolean operators, while the leaves represent either seed functions or additional independent variables. The fitness functions evaluate the generality of construction by measuring the balancedness and the nonlinearity of the resulting Boolean functions starting from a set of optimal seeds. Our experiments show that, for certain parameters combinations, GP always converges to a general construction. A closer inspection of these solutions reveals that GP actually finds many equivalent constructions, and the solutions that we analyzed in detail turned out to be a particular case of the indirect sum construction~\cite{carlet_2021}.

Our findings seem to indicate that GP cannot find novel constructions with our current formulation of the optimization problem. However, it is still remarkable that GP always finds the same simple construction in all considered experiments, albeit under different syntactic forms. One possible explanation for this behavior could be related to the genotype representation adopted in our experiments. Indeed, the additional two variables are always used \emph{externally to the seed functions}; in other words, $v_0$ and $v_1$ are never employed as inputs to the seed functions themselves, but rather their values are combined with the outputs of the seeds. Remark this is not a real restriction from the semantic point of view, since all Boolean functions of $n+2$ variables can be expressed as the combination of two $n$-variable functions $f,g$ with the additional two variables; However, considering also that we enforce a maximum depth on the trees, the representations that GP can evolve in this way are quite constrained. In particular, we formulate the hypothesis that the concatenation construction is the \emph{only} general construction discoverable by GP under this encoding and that differences arise only at a syntactic level, with more or less bloated constructions. We plan to investigate this hypothesis in future research, following two complementary future directions. The first direction is to investigate if, under the given encoding constraints, the concatenation is the only optimal solution in the \emph{semantic space} of constructions. This could be accomplished by analyzing the space of all constructions in terms of their truth tables. It would also be interesting to consider the use of geometric semantic GP~\cite{moraglio12} for this particular problem. Finally, the second direction is to experiment with GP encodings that are less constrained, by either allowing the additional variables to partake in the input of the seed functions, or by adopting a more general approach with independent additional variables. Indeed, the indirect sum construction is more symmetric in its structure than what we experimented with in this paper, and this the reason why GP could not evolve it in its most general form.

\bibliographystyle{abbrv}
\bibliography{references}

\end{document}